\documentclass[runningheads]{llncs}
\usepackage[colorlinks=true,allcolors=blue]{hyperref}
\usepackage{amsmath,amssymb,amsfonts}

\usepackage{multirow}
\usepackage{booktabs}
\usepackage{placeins}
\usepackage{bm}
\usepackage{subfigure} 
\usepackage{floatrow}
\newfloatcommand{capbtabbox}{table}[][\FBwidth]
\usepackage{graphicx}
\usepackage[misc]{ifsym}
\begin{document}
\title{Zero-shot Nuclei Detection via Visual-Language Pre-trained Models}
%
%
\author{Yongjian Wu\inst{1}\orcidID{0009-0000-5631-164X} \and
Yang Zhou\inst{1}\orcidID{0000-0003-2848-7642} \and
Jiya Saiyin\inst{1} \and
Bingzheng Wei\inst{2}\orcidID{0000-0001-6979-0459} \and
Maode Lai\inst{3} \and
Jianzhong Shou\inst{4} \and
Yubo Fan\inst{1}\orcidID{0000-0002-3480-4395} \and 
Yan Xu\inst{1}$^{(\textrm{\Letter})}$\orcidID{0000-0002-2636-7594}}
%
\authorrunning{Y. Wu, Y. Zhou et al.}
%
\institute{School of Biological Science and Medical Engineering, State Key Laboratory of Software Development Environment, Key Laboratory of Biomechanics and Mechanobiology of Ministry of Education, Beijing Advanced Innovation Center for Biomedical Engineering, Beihang University, Beijing 100191, China \email{xuyan04@gmail.com} \and
Xiaomi Corporation, Beijing 100085, China \and
Department of Pathology, School of Medicine, Zhejiang University, Zhejiang Provincial Key Laboratory of Disease Proteomics and Alibaba-Zhejiang University Joint Research Center of Future Digital Healthcare, Hangzhou 310053, China \and
Chinese Academy of Medical Sciences and Peking Union Medical College, Beijing 100021, China}
%
\maketitle              
\begin{abstract}
Large-scale visual-language pre-trained models (VLPM) have proven their excellent performance in downstream object detection for natural scenes. However, zero-shot nuclei detection on H\&E images via VLPMs remains underexplored. The large gap between medical images and the web-originated text-image pairs used for pre-training makes it a challenging task. In this paper, we attempt to explore the potential of the object-level VLPM, Grounded Language-Image Pre-training (GLIP) model, for zero-shot nuclei detection. Concretely, an automatic prompts design pipeline is devised based on the association binding trait of VLPM and the image-to-text VLPM BLIP, avoiding empirical manual prompts engineering. We further establish a self-training framework, using the automatically designed prompts to generate the preliminary results as pseudo labels from GLIP and refine the predicted boxes in an iterative manner. Our method achieves a remarkable performance for label-free nuclei detection, surpassing other comparison methods. Foremost, our work demonstrates that the VLPM pre-trained on natural image-text pairs exhibits astonishing potential for downstream tasks in the medical field as well. Code will be released at \href{https://github.com/wuyongjianCODE/VLPMNuD}{github.com/VLPMNuD}.
\keywords{ Nuclei Detection \and Unsupervised Learning \and Visual-Language Pre-trained Models \and Prompt Designing \and Zero-shot Learning.}
\end{abstract}
\footnotetext[1]{Equal contribution: Yongjian Wu, Yang Zhou. Corresponding author: Yan Xu.}
%
%
\section{Introduction}

In the field of medical image processing, nuclei detection on Hematoxylin and Eosin (H\&E)-stained images plays a crucial role in various areas of biomedical research and clinical applications \cite{gleason1992histologic}. While fully-supervised methods have been proposed for this task \cite{mahanta2021ihc,graham2019hover,yi2019multi}, the annotation remains labor-intensive and expensive. To address the aforementioned issue, several unsupervised methods have been proposed, including thresh-holding-based methods \cite{jiao2006improved,mouelhi2018fast}, self-supervised-based methods \cite{sahasrabudhe2020self}, and domain adaptation-based methods \cite{le2022unsupervised}. Among these, domain adaptation methods are mainstream and have demonstrated favorable performance by achieving adaptation through aligning the source and target domains \cite{le2022unsupervised}. However, current unsupervised methods exhibit strong empirical design and introduce subjective biases during the model design process, thus current unsupervised methods may lead to suboptimal results.

Yet, newly developed large-scale visual-language pre-trained models (VLPMs) have provided another possible unsupervised learning paradigm \cite{radford2021learning,jia2021scaling}. VLPM learns aligned text and image features from massive text-image pairs acquired from the internet, making the learned visual features semantic-rich, general, and transferable. Zero-shot learning methods based on VLPM for downstream tasks such as text-driven image manipulation \cite{patashnik2021styleclip}, image captioning \cite{li2022blip}, view synthesis \cite{jain2021putting}, and object detection \cite{li2022grounded}, have achieved excellent results. 

Among VLPMs, Grounded Language-Image Pre-training (GLIP) model \cite{li2022grounded}, pre-trained at the object level, can even rival fully-supervised counterparts in zero-shot object detection and phrase grounding tasks. Although VLPM has been utilized for object detection in natural scenes, zero-shot nuclei detection on H\&E images via VLPM remains underexplored. The significant domain differences between medical H\&E images and the natural images used for pre-training make this task challenging. It is wondered whether VLPM, with its rich semantic information, can facilitate direct prediction of nuclei detection through semantic-driven prompts, establishing an elegant, concise, clear but more efficient and transferable unsupervised system for label-free nuclei detection.

Building upon this concept, our goal is to establish a zero-shot nuclei detection framework based on VLPM. However, directly applying VLPM for this task poses two challenges. (1) Due to the gap between medical images and the web-originated text-image pairs used for pre-training, the text-encoder may lack prior knowledge of medical concept words, thus making the prompt design for zero-shot detection a challenging task. (2) Different from the objects in natural images, the high density and specialized morphology of nuclei in H\&E stained images may lead to missed detection, false detection, and overlapping during zero-shot transfer solely with prompts.

To address the first challenge, Yamada \textit{et al.} have analyzed and revealed that under the pre-training of vast text-image pairs, VLPM establishes a strong association binding between the object and its semantic attributes regardless of image domain, i.e., associated attribute text can fully describe the corresponding objects in an image through VLPM \cite{yamada2022lemons}. Therefore, it is feasible for VLPM to detect unseen medical objects in a label-free manner by constructing appropriate attribute texts. Manual prompting is a cumbersome and subjective process, which may lead to considerable bias. Yet, the VLPM network BLIP \cite{li2022blip} has the capability to generate automatic descriptions for images. Therefore, we first use BLIP to automatically generate attribute words to describe the unseen nuclei object. This approach avoids the empirical manual prompt design and fully leverages the text-to-image aligning trait of VLPMs. We subsequently integrate these attribute words with medical nouns, i.e. ``[shape][color][noun]'', to create detection prompts. These prompts are then inputted into GLIP to realize zero-shot detection of nuclei. Our proposed automatic prompt designing method fully utilizes the text-to-image alignment of VLPM, and enables the automatic generation of the most suitable attribute text words describing the corresponding domain. Our approach offers excellent interpretability.

Through GLIP's strong object retrieval performance, we can obtain preliminary boxes. The precision of these preliminary boxes is relatively high, but there is still considerable room for improvement in recall. Therefore, we further establish a self-training framework. We use the preliminary boxes generated by GLIP as pseudo labels for further training YOLOX \cite{ge2021yolox}, to refine and polish the predicted boxes in an iterative manner. Together with the self-training strategy, the resulting model achieves a remarkable performance for label-free nuclei detection, surpassing other comparison methods. We demonstrate that VLPM, which is pre-trained on natural image-text pairs, also exhibits astonishing potential for downstream tasks in the medical field.

The contributions of this paper are threefold. (1) A novel zero-shot label-free nuclei detection framework is proposed based on VLPMs. Our method outperforms all existing unsupervised methods and demonstrates excellent transferability. (2) We leverage GLIP, which places more emphasis on object-level representation learning and generates more high-quality language-aware visual representations compared to Contrastive Language-Image Pre-training (CLIP) model, to achieve better nuclei retrieval. (3) An automatic prompt design process is established based on the association binding trait of VLPM to avoid non-trivial empirical manual prompt engineering.

\section{Method}

Our approach aims to establish a zero-shot nuclei detection framework based on VLPMs by directly using text prompts. We utilize GLIP for better object-level representation extraction. The overview of our framework is shown in Fig.\ref{fig1}.

\subsection{Object-level VLPM \textemdash GLIP}

\begin{figure}[t]
\includegraphics[width=\textwidth]{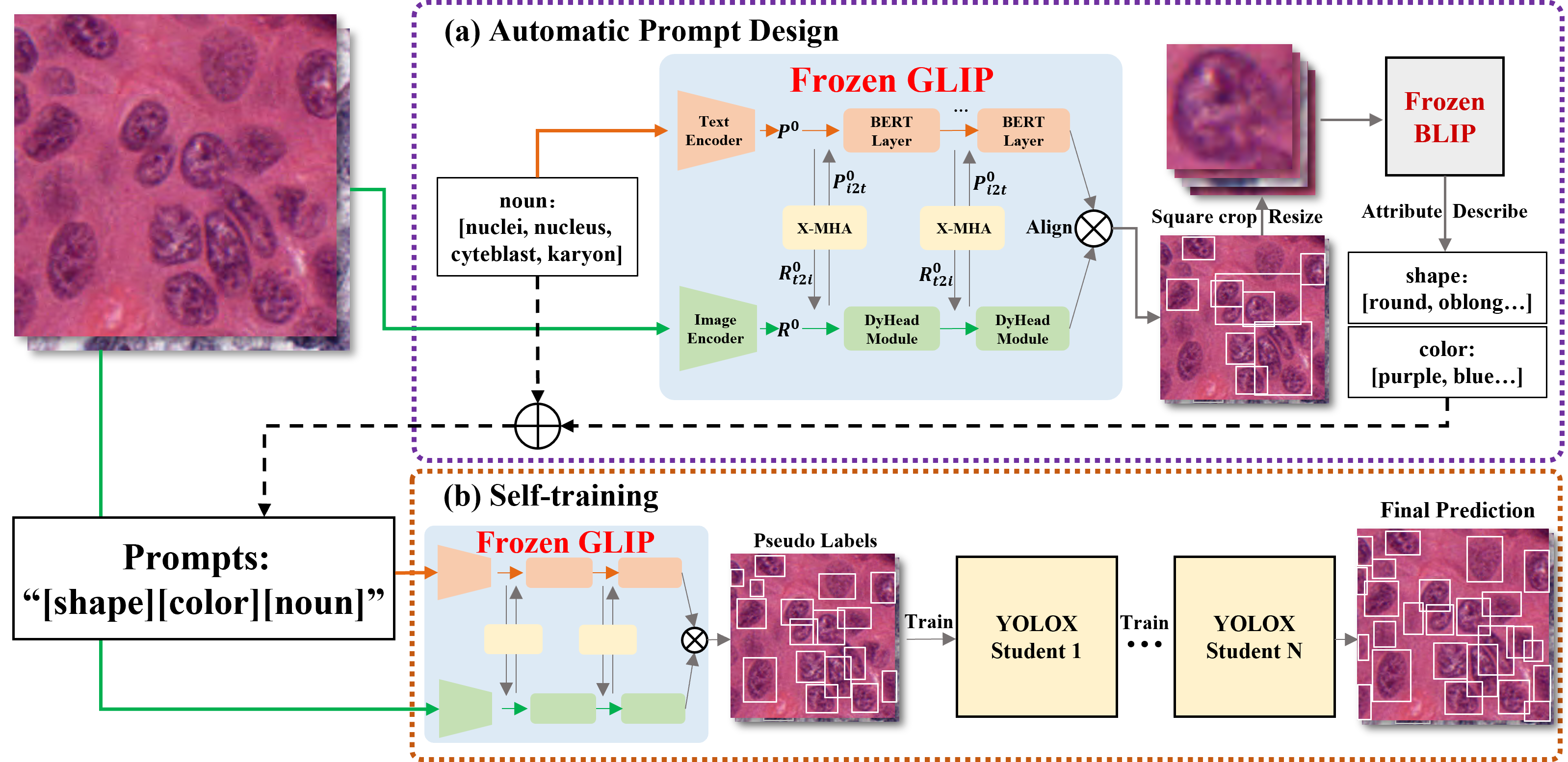}
\caption{An overview of our zero-shot nuclei detection framework based on the frozen object-level VLPM GLIP. (a) Given the original target nouns of the task, prompts are designed automatically to avoid non-trivial empirical manual prompt engineering, based on the association binding of VLPM and the image-to-text VLPM BLIP. (b) A self-training strategy is further adopted to refine and polish the predicted boxes in an iterative manner. The automatically designed prompts are used by GLIP to generate the preliminary results as pseudo labels.} \label{fig1}
\end{figure}

Recently, large VLPMs such as CLIP \cite{radford2021learning} and ALIGN \cite{jia2021scaling} have made great progress in generic visual representation learning and demonstrated the enormous potential of utilizing prompts for zero-shot transfer. 

A typical VLPM framework comprises two encoders: the text encoder, which encodes text prompts to semantic-rich text embeddings, and the image encoder, which encodes images to visual embeddings. These VLPMs use a vast amount of web-originated text-image pairs $\{ (X,T )\}_{i}$ to learn the text-to-image alignment through contrastive loss over text and visual embeddings. Denoting the image encoder as $E_{I}$, the text encoder as $E_{T}$, the cosine similarity function as $\cos(\cdot)$, and assuming that $K$ text-image pairs are used in each training epoch, the objective of the contrastive learning can be formulated as:
\begin{equation}
\label{eq1}
\mathcal{L}_c=-\sum_{i=1}^K \log \frac{\exp \left(\cos \left(\cdot E_I\left(X_i\right) \cdot E_T\left(T_i\right)\right) / \tau\right)}{\sum_{j=1}^K \exp \left(\cos \left(E_I\left(X_i\right) \cdot E_T\left(T_j\right)\right) / \tau\right)}.
\end{equation}
Through this aligning contrastive learning, VLPM aligns text and image in a common feature space, allowing one to directly transfer a trained VLPM to downstream tasks via manipulating text, i.e., prompt engineering. The visual representations of input images are semantic-rich and interpretability-friendly with the help of aligned text-image pairs.

However, the conventional VLPM pre-training process only aligns the image with the text from a whole perspective, which results in a lack of emphasis on object-level representations. Therefore, Li \textit{et al.} proposed GLIP \cite{li2022grounded}, whose image encoder generates visual embeddings for each object of different regions in the image to align with object words present in the text. Moreover, GLIP utilizes web-originated phrase grounding text-image pairs to extract novel object word entities, expanding the object concept in the text encoder. Additionally, unlike CLIP, which only aligns embeddings at the end of the model, GLIP builds a deep cross-modality fusion based on cross-attention \cite{chen2021crossvit} for multi-level alignment. As shown in Fig.\ref{fig1}(a), GLIP leverages DyheadModules \cite{dai2021dynamic} and BERTLayer \cite{devlin2018bert} as the image and text encoding layers, respectively. With text embedding represented as $R$ and visual embedding as $P$, the deep fusion process can be represented as:
\begin{equation}
\label{eq2}
R_{\mathrm{t} 2 \mathrm{i}}^i, P_{\mathrm{i} 2 \mathrm{t}}^i=\textit{X-MHA}\left(R^i, P^i\right), \cdots i \in\{0,1, \ldots, L-1\},
\end{equation}
\begin{equation}
\label{eq3}
R^{i+1}=\textit{DyHeadModule}\left(R^i+R_{t 2 \mathrm{i}}^i\right), \cdots R=R^L,
\end{equation}
\begin{equation}
\label{eq4}
P^{i+1}=\textit{BERTLayer}\left(P^i+P_{t 2 \mathrm{i}}^i\right), \cdots P=P^L,
\end{equation}
where $L$ is the total number of layers, $R^0$ denotes the visual features from swin-transformer-large \cite{liu2021swin}, and $P^0$ denotes the token features from BERT \cite{devlin2018bert}. \textit{X-MHA} represents cross-attention. This architecture enables GLIP to attain superior object-level performance and semantic aggregation. Consequently, GLIP is better suited for object-level zero-shot transfer than conventional VLPMs. Thus, we adopt GLIP to extract better object-level representations for nuclei detection.
\subsection{Automatic Prompt Design}
The text input, known as the prompt, plays a crucial role in the zero-shot transfer of VLPM for downstream tasks. GLIP originally uses concatenated object nouns such as "object noun 1. Object noun 2..." as default text prompts for detection, and also allows for manual engineering to improve performance \cite{li2022grounded}. However, manual prompt engineering is a non-trivial challenge, demanding substantial effort and expertise. Furthermore, a notable disparity exists between the web-originated pre-training text-image pairs and medical images. Thus simple noun concatenation is insufficient for GLIP to retrieve nuclei.

We note that Yamada \textit{et al.} prove that the pre-training on extensive text-image pairs has allowed VLPMs to establish a strong association binding between objects and their semantic attributes \cite{yamada2022lemons}. Thus, through VLPM, the associated attribute text can accurately depict the corresponding object. Based on this research, we propose that VLPM has the potential to detect unlabelled medical objects in a zero-shot manner by constructing relevant attribute text.

As shown in Fig.\ref{fig1}(a), we first use the image captioning VLPM BLIP \cite{li2022blip} to automatically generate attribute words to describe the unseen nuclei object. BLIP allows us to avoid manual attribute design and generates attribute vocabulary that conforms to the text-to-image alignment of VLPM. This process involves three steps. (1) Directly input target medical nouns into GLIP for coarse box prediction. (2) Use the coarse boxes to squarely crop the image, the cropped objects are fed into a frozen BLIP to automatically generate attribute words that describe the object. (3) Word frequency statistics and classification are adopted to find the top $M$ words that describe the object's shape and color, respectively, for that ``shape'' and ``color'' are the most relative two attributes that depict nuclei. For a thorough description, we augment the attribute words with synonyms retrieved by a pre-trained language model \cite{radford2019language}. All these attribute words are combined with those medical nouns to automatically generate a triplet detection prompt of ``[shape][color][noun]''. Finally, all generated triplets were put into GLIP for refined detection. This method avoids the empirical manual prompt design and fully utilizes the text-to-image aligning trait of VLPMs.

Our automatic prompt design leverages the powerful text-to-image aligning capabilities of GLIP and BLIP. This approach also enables the automatic generation of the most appropriate attribute words for the specific domain, embodying excellent interpretability.
\subsection{Self-training Boosting}
Leveraging the strong object retrieval performance of GLIP, we obtain preliminary detection boxes with high precision but low recall. These boxes suffer from missed detection, false detection, and overlapping. To fully exploit the zero-shot potential of GLIP, a self-training framework is established. The automatic prompts are inputted into GLIP to generate the initial results which served as pseudo labels for training YOLOX \cite{ge2021yolox}. Then, the converged YOLOX is used as a teacher to generate new pseudo labels, and iteratively trains students YOLOX. As self-training is based on the EM optimization algorithm \cite{moon1996expectation}, it propels our system to continuously refine the predicted boxes and achieve a better optimum. 

\section{Experiments}
\subsection{Dataset and Implementation}
The dataset used in this study is the MoNuSeg dataset \cite{kumar2017dataset}, which consists of 30 nuclei images of size $1000\times1000$, with an average of 658 nuclei per image. Following Kumar \textit{et al.} \cite{kumar2017dataset}, the dataset was split into training and testing sets with a ratio of 16:14. 16 training images served as inputs of GLIP to generate pseudo-labels for self-training, with 4 images randomly selected for validation. Annotations were solely employed for evaluation purposes on the test images. 16 overlapped image patches of size $256\times256$ were extracted from each image and randomly cropped into $224\times224$ as inputs.

In terms of experimental settings, four Nvidia RTX 3090 GPUs were utilized, each with 24 GB of memory. In the automatic prompt generating process, the VQA weights of BLIP finetuned on ViT-B and CapFilt-L \cite{li2022blip} were used to generate [shape] and [color] attributes. These attribute words are augmented with synonyms by GPT \cite{radford2019language}, i.e. attribute augmentation. The target medical noun list was first set to [``nuclei''] straightforwardly, and was also augmented by GPT to [``nuclei'', ``nucleus'', ``cyteblast'', ``karyon''], i.e. noun augmentation. Attribute words were subsequently combined with the target medical nouns  to ``[shape][color][noun]'' format as inputs of GLIP to generate bounding boxes as pseudo labels. The weights used for GLIP is GLIP-L. For self-training refinement, we used the default setting of YOLOX and followed the standard self-training methodology described in \cite{dopido2013semisupervised}.
\subsection{Comparison}
Our proposed method was compared with the representative fully-supervised method YOLOX \cite{ge2021yolox}, as well as the current state-of-the-art (SOTA) methods in unsupervised object detection, including SSNS \cite{sahasrabudhe2020self}, SOP \cite{le2022unsupervised}, and VLDet \cite{lin2022learning}. Among them, the fully-supervised method YOLOX represents the current SOTA on natural images, and VLDet is a newly proposed zero-shot object detection method based on CLIP. For evaluation, mAP, AP50, AP75, and AR are chosen as metrics, following COCO \cite{lin2014microsoft}. The final results are shown in Table \ref{tab：comparison}.

Referring to the table, it is evident that our GLIP-based approach outperforms all unsupervised techniques, including domain adaptation-based and clip-based methods. Fig.\ref{fig2} depicts the visualization of the detection results.
\begin{table}[t]
\caption{Comparison results on MoNuSeg \cite{kumar2017dataset}. The best results of unsupervised methods are marked in \textbf{bold}.}
\begin{center}
\begin{tabular}{c|c|cccc}
\hline
\textbf{supervision} & \textbf{method} & \textbf{mAP} & \textbf{AP50} & \textbf{AP75} & \textbf{AR} \\ \hline
\textbf{fully-supervised} & \textbf{YOLOX (2021) \cite{ge2021yolox}} & 0.447 & 0.832 & 0.437 & 0.528 \\ \hline
\multirow{4}{*}{\textbf{unsupervised}} & \textbf{SSNS (2020) \cite{sahasrabudhe2020self}} & 0.354 & 0.739 & 0.288 & 0.441 \\
 & \textbf{SOP (2022) \cite{le2022unsupervised}} & 0.235 & 0.609 & 0.096 & 0.351 \\
 & \textbf{VLDet (2023) \cite{lin2022learning}} & 0.173 & 0.407 & 0.112 & 0.263 \\
 & \textbf{Ours} & \textbf{0.416} & \textbf{0.808} & \textbf{0.382} & \textbf{0.502} \\ \hline
\end{tabular}
\label{tab：comparison}
\end{center}
\end{table}
\begin{figure}[t]
\includegraphics[width=\textwidth]{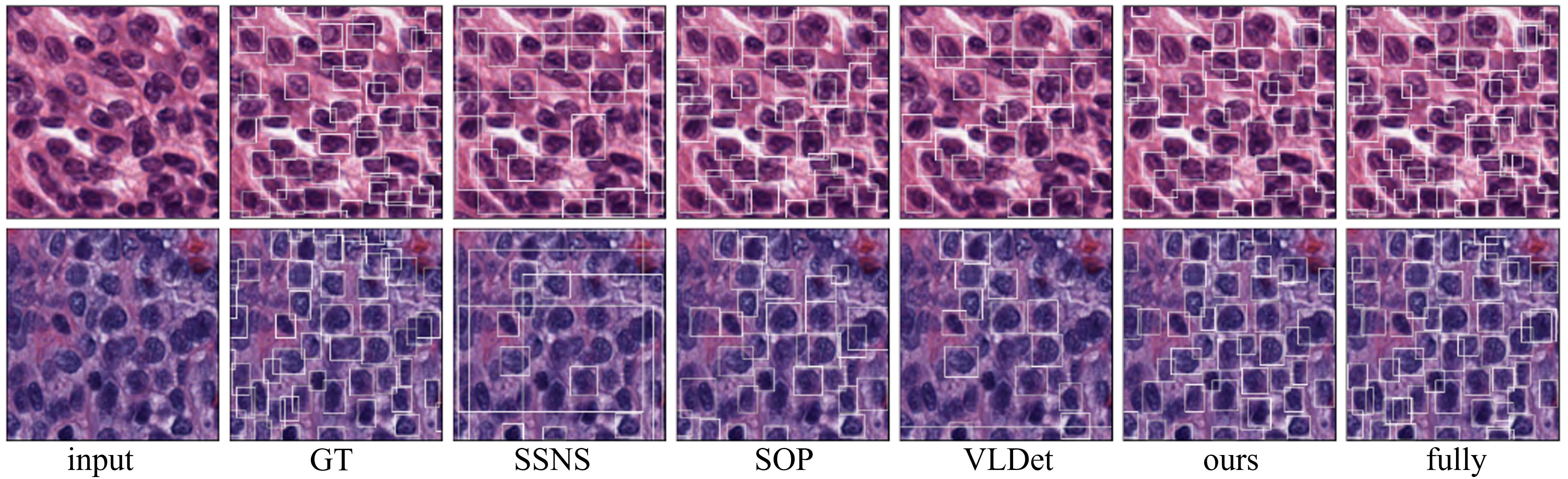}
\caption{Output visualizations of different models. The boxes are shown in white.} \label{fig2}
\end{figure}
\subsection{Ablation Studies}
\subsubsection{Automatic Prompt Design}
Firstly, we conducted an ablation study specifically targeting the prompt while ensuring that other conditions remained constant, the results are presented in Table \ref{tab:prompt}. 

\begin{table*}[t]
\begin{floatrow}
\capbtabbox{ 
\resizebox{0.45\textwidth}{!}{
\renewcommand{\arraystretch}{1}
\begin{tabular}{c|cccc}
\hline
\textbf{prompt design} & \textbf{mAP} & \textbf{AP50} & \textbf{AP75} & \textbf{AR} \\ \hline
\textbf{noun. \cite{li2022grounded}} & 0.064 & 0.152 & 0.036 & 0.150 \\
\textbf{manual design \cite{li2022grounded}} & 0.414 & 0.757 & 0.422 & 0.509 \\ \hline
\textbf{auto: {[}shape{]}{[}noun.{]}} & 0.336 & 0.604 & 0.350 & 0.434 \\
\textbf{auto: {[}color{]}{[}noun.{]}} & 0.213 & 0.447 & 0.176 & 0.325 \\
\textbf{auto: {[}shape{]}{[}color{]}} & 0.413 & 0.726 & 0.438 & 0.498 \\
\textbf{auto:{[}shape{]}{[}color{]}{[}noun.{]}} & \textbf{0.416} & \textbf{0.808} & \textbf{0.382} & \textbf{0.502} \\ \hline
\end{tabular} 
}
}{
 \caption{Results of adopting different prompt design methods. The pompts of last 4 rows are automatically generated. The best results are marked in \textbf{bold}.}
 \label{tab:prompt}
}

\capbtabbox{ 
\resizebox{0.45\textwidth}{!}{
\renewcommand{\arraystretch}{1.13}
\begin{tabular}{c|c|cccc}
\hline
\textbf{A Aug.} & \textbf{N Aug.} & \textbf{mAP} & \textbf{AP50} & \textbf{AP75} & \textbf{AR} \\ \hline
 &  & 0.372 & 0.725 & 0.337 & 0.464 \\
\checkmark &  & \textbf{0.416} & \textbf{0.808} & \textbf{0.382} & \textbf{0.502} \\
 & \checkmark & 0.316 & 0.754 & 0.182 & 0.419 \\
\checkmark & \checkmark & 0.332 & 0.659 & 0.296 & 0.434 \\ \hline
\end{tabular}
}
}{
 \caption{Ablation study on word augmentation. ``A Aug.'' and ``N Aug.'' refer to attribute augmentation and noun augmentation, respectively.}
 \label{tab: aug}
}

\end{floatrow} 
\end{table*}


The first row of the table displays the default noun-concatenation prompt GLIP originally adopted, i.e. ``nuclei. nucleus. cyteblast. karyon''. The second row represents the same set of nouns with some manual property descriptions added, like ``Nuclei. Nucleus. cyteblast. karyon, which are round or oval, and purple or magenta''. It is noteworthy that this manual approach is empirically subjective and therefore prone to significant biases. The subsequent rows in the table demonstrate the combination of attributes generated by our automatic prompt design method. In these experiments, $M$ was set to 3. 

It is worth noting that the predictions generated by the prompts shown in the first and second rows also employed the self-training strategy until convergence. However, the results of the first row contain a significant amount of noise, implying that the intrinsic gap between medical nouns and natural nouns impedes the directly zero-shot transfer of GLIP. The second row improves obviously, indicating the effectiveness of attribute description. But manual design is empirical and tedious. As for the automatically generated prompts, it is evident that as the description of attributes becomes comprehensive, from only including shape or color solely to encompassing both, GLIP's performance improves gradually. Furthermore, the second-to-last row indicates that even without nouns, attribute words alone can achieve good results, which also demonstrates the ability of BLIP-generated attribute words to effectively describe the target nuclei. Through the utilization of VLPM's text-to-image alignment capabilities, the proposed automatic prompt design method generates the most suitable attribute words for a given domain automatically, with a high degree of interpretability. Please refer to the supplement for a detailed list of [shape] and [color] attributes.

We further looked into the effect of word augmentation. The results are shown in Table \ref{tab: aug}. Without and with noun augmentation, the noun lists were [``nuclei''] and [``nuclei'', ``nucleus'', ``cyteblast'', ``karyon''], respectively. The first row of Table \ref{tab: aug} uses non-augmented ``[shape][color][noun]'', while the first row of Table \ref{tab:prompt} uses noun-augmented concatenation. It is intriguing that applying noun augmentation may lead to suboptimum results compared with the counterparts using the straightforward [``nuclei'']. This is most likely because the augmented new synonym nouns are uncommon medical words and did not appear in the pre-training data that GLIP used. However, applying attribute word augmentation is generally effective because augmented attribute words are also common descriptions for natural scenes. These results suggest a general approach for leveraging the VLPM's potential in downstream medical tasks, that is identifying common attribute words that can be used for describing the target nouns.

\subsubsection{Self-training and YOLOX}
The box optimization process of the self-training stage is recorded, and the corresponding results are presented in the supplement. YOLOX and self-training are not the essential reasons for the superior performance of our method. The true key is the utilization of semantic-information-rich VLPMs. To illustrate this point, we employed another commonly used unsupervised detection method, superpixels \cite{achanta2012slic}, to generate pseudo labels in a zero-shot manner for a fair comparison. These pseudo labels were then fed into the self-training framework based on the YOLOX segmentation architecture, keeping the settings consistent with our approach except for the pseudo label generation. Additionally, we also used DETR \cite{carion2020end} instead of YOLOX in our method. The results are shown in the supplement and demonstrate that the high performance of our method lies in the effective utilization of the knowledge from VLPMs rather than YOLOX or self-training.




\section{Conclusion}

In this work, we propose to use the object-level VLPM, GLIP, to realize zero-shot nuclei detection on H\&E images. An automatic prompt design pipeline is proposed to avoid empirical manual prompt design. It fully utilizes the text-to-image alignment of BLIP and GLIP, and enables the automatic generation of the most suitable attribute describing words, offering excellent interpretability. Furthermore, we utilize the self-training strategy to polish the predicted boxes in an iterative manner. Our method achieves a remarkable performance for label-free nuclei detection, surpassing other comparison methods. We demonstrate that VLPMs pre-trained on natural image-text pairs still exhibit astonishing potential for downstream tasks in the medical field.

%
%


\bibliographystyle{splncs04}
\bibliography{mybibliography}
\newpage

\section*{Supplement}
\setcounter{table}{0}
\setcounter{figure}{0}
\begin{table}[h]
\caption{Effectiveness of self-training strategy. T refers to the teacher and S refers to the students. The results shown here are calculated on the testing set, while the pseudo labels are generated from the training set. The precision and recall are calculated using a threshold IoU of 0.5. The best results are marked in bold. The initial boxes precision obtained by GLIP is relatively high, but there is still a lot of room for improvement in recall. Through self-training, the overall performance continually improved until convergence.}
\begin{center}
\begin{tabular}{c|cccccc}
\hline
\textbf{Teacher/Students} & \textbf{Precision} & \textbf{Recall} & \textbf{mAP} & \textbf{AP50} & \textbf{AP75} & \textbf{AR} \\ \hline
\textbf{Raw GLIP T0} & 0.678          & 0.310          & 0.146          & 0.252          & 0.165          & 0.213          \\
\textbf{YOLOX S1}    & 0.733          & 0.706         & 0.293          & 0.636          & 0.224          & 0.397          \\
\textbf{YOLOX S2}    & 0.807          & 0.754         & 0.359          & 0.698          & 0.335          & 0.445          \\
\textbf{YOLOX S3}    & \textbf{0.814} & 0.815         & 0.404          & 0.755          & 0.395          & 0.497          \\
\textbf{YOLOX S4}    & 0.813          & \textbf{0.820} & \textbf{0.416} & \textbf{0.808} & 0.382          & 0.502          \\
\textbf{YOLOX S5}    & \textbf{0.814} & 0.807         & 0.414          & 0.742          & \textbf{0.436} & \textbf{0.509} \\ \hline
\end{tabular}
\vskip -0.45in
\label{tab：self}
\end{center}
\end{table}
\begin{table}[h]
\caption{Illustration of the significance of the automatic prompt design using VLPMs for pseudo label generation. Another commonly used unsupervised detection method, superpixels (\textit{Achanta, Radhakrishna, et al. SLIC superpixels compared to state-of-the-art superpixel methods[J]//IEEE TPAMI 2012}), is used to generate pseudo labels in a zero-shot manner. These pseudo labels were then fed into the self-training framework based on the YOLOX segmentation architecture, keeping the settings consistent with using VLPMs for pseudo label generation. The results reveal poor performance.}
\begin{center}
\begin{tabular}{c|cccc}
\hline
\textbf{Teacher/Students} & \textbf{mAP} & \textbf{AP50} & \textbf{AP75} & \textbf{AR} \\ \hline
\textbf{Superpixels T0} & 0.027 & 0.075 & 0.012 & 0.035 \\
\textbf{YOLOX s1} & 0.260 & 0.612 & 0.162 & 0.373 \\
\textbf{YOLOX s2} & 0.272 & 0.617 & 0.183 & 0.392 \\
\textbf{YOLOX s3} & \textbf{0.284} & \textbf{0.655} & 0.169 & \textbf{0.404} \\
\textbf{YOLOX s4} & 0.279 & 0.614 & \textbf{0.213} & 0.389 \\ \hline
\end{tabular}
\vskip -0.45in
\label{tab：superpixel}
\end{center}
\end{table}
\begin{table}[h]
\caption{Results of using DETR (\textit{Carion, Nicolas, et al. End-to-end object detection with transformers[C]//Computer Vision–ECCV 2020}) instead of YOLOX as the detection architecture for self-training in our method.}
\begin{center}
\begin{tabular}{c|cccc}
\hline
\textbf{method} & \textbf{mAP} & \textbf{AP50} & \textbf{AP75} & \textbf{AR} \\ \hline
\textbf{fully-supervised DETR} & 0.404 & 0.749 & 0.398 & 0.501 \\
\textbf{Ours+DETR} & 0.388 & 0.731 & 0.376 & 0.487 \\ \hline
\end{tabular}
\vskip -0.45in
\label{tab：detr}
\end{center}
\end{table}

\begin{figure}[h]
\includegraphics[width=\textwidth]{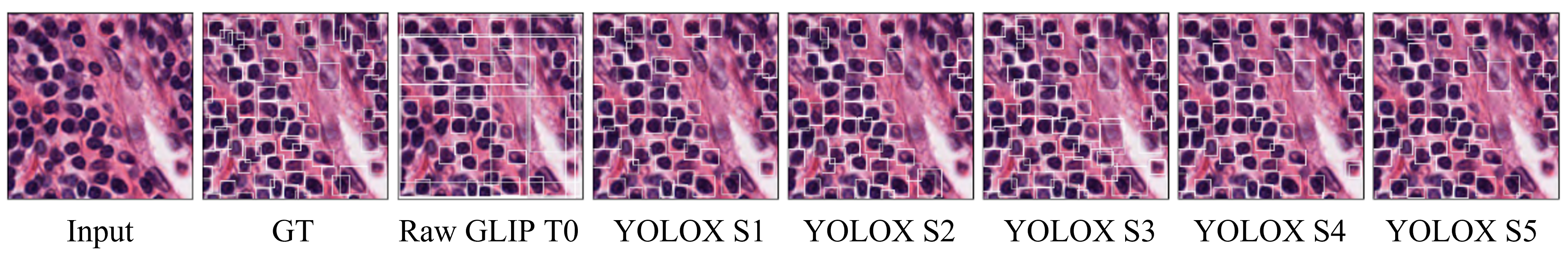}
\caption{Output visualizations of different stages in the self-training process. From left to right: input image, the ground truth, raw GLIP teacher, YOLOX students.} \label{fig2}
\end{figure}
\begin{table}[t]
\caption{Detailed list of [shape] and [color]. The first row contains the 3 most frequent words generated by BLIP. The second row shows the synonyms augmented by GPT.}
\begin{center}
\resizebox{\textwidth}{!}{
\begin{tabular}{c|c|c}
\hline
\textbf{}          & \textbf{{[}shape{]}}                       & \textbf{{[}color{]}}                            \\ \hline
\textbf{From BLIP} & round, circle, oblong                      & blue, black, purple                               \\ \hline
\multirow{3}{*}{\textbf{Synonyms}} & oval, elliptical, rectangular, Oblate, & magenta, deep purple, dark purple, lilac, \\
                   & rectilinear, circular, ring, disc, loop, cycle, & violet, azure, aqua, cyan, light blue, turquoise, \\
                   & spherical, curved, rounded, cyclical & light blue, sky blue, ebony, coal, jet, pitch, onyx \\ \hline
\end{tabular}%
}
\label{tab：words}
\end{center}
\end{table}
\begin{table}[t]
\caption{Comparison with CLIP-based zero-shot object detection methods, VL-PLM (\textit{Zhao S, Zhang Z, Schulter S, et al. Exploiting unlabeled data with vision and language models for object detection[C]//Computer Vision–ECCV 2022}) and VLDet (\textit{Lin C, Sun P, Jiang Y, et al. Learning Object-Language Alignments for Open-Vocabulary Object Detection[C]//ICLR 2023}). Our method is GLIP-based. The results demonstrate the superiority of GLIP over CLIP in zero-shot transfer to nuclei detection.}
\centering
\begin{tabular}{c|cccc}
\hline
\textbf{method}        & \textbf{mAP}   & \textbf{AP50}  & \textbf{AP75}  & \textbf{AR}    \\ \hline
\textbf{VL-PLM (2022)} & 0.333          & 0.582          & 0.342          & 0.501          \\
\textbf{VLDet (2023)}  & 0.173          & 0.407          & 0.112          & 0.263          \\ \hline
\textbf{Ours}          & \textbf{0.416} & \textbf{0.808} & \textbf{0.382} & \textbf{0.502} \\ \hline
\end{tabular}%
\end{table}
%




\end{document}